\documentclass[letterpaper, 10pt, conference]{ieeeconf}
\IEEEoverridecommandlockouts

\usepackage[pdftex]{graphicx}
\usepackage{multicol}
\usepackage{multirow}
\usepackage{amsmath, amsfonts, amssymb}
\usepackage{xspace}
\usepackage{amsmath}
\usepackage{latexsym}
\usepackage{url}
\usepackage{listings}
\usepackage{float}
\usepackage{tabularx}
\usepackage{tikz}
\usepackage{fancybox}
\usepackage{flushend}
\usepackage{blindtext}
\usepackage{tabulary}
\usepackage{color, colortbl}

\usepackage[ruled,lined]{algorithm2e}

\lstdefinelanguage[OWL]{XML} 
{morekeywords={Individual,ObjectProperty,Types,Facts,Class,SubClassOf,Domain,
Range,SubPropertyOf,EquivalentTo}}
\lstdefinestyle{OWL}    {language=[OWL]XML, lineskip=-0.3ex, fontadjust=true, 
basicstyle={\scriptsize\nopagebreak[4]}}
\lstdefinestyle{Prolog} {language=Prolog,   lineskip=-0.3ex, fontadjust=true, 
basicstyle={\scriptsize\nopagebreak[4]}, commentstyle=\scriptsize, 
showstringspaces=false, showspaces=false, showtabs=false}

\definecolor{Gray}{gray}{0.5}
\definecolor{LightCyan}{rgb}{0.88,1,1}
\graphicspath{{../../../img/}}

\title{Amortized Object and Scene Perception \\for Long-term 
Robot Manipulation}

\author{Ferenc B\'alint-Bencz\'edi and Michael Beetz
\thanks{Authors are with the Institute for Artificial Intelligence, 
University of Bremen}
\thanks{ \tt\small \{balintbe,beetz\}@cs.uni-bremen.de}
}

\usetikzlibrary{arrows}

\tikzset{
  treenode/.style = {align=center, inner sep=0pt, text centered,
    font=\sffamily},
  arn_n/.style = {treenode, circle, white, font=\sffamily\bfseries, draw=black,
    fill=black, text width=1.5em},
  arn_r/.style = {treenode, circle, red, draw=red, 
    text width=1.5em, very thick},
  arn_x/.style = {treenode, rectangle, draw=black,
    minimum width=0.5em, minimum height=0.5em}
}

\begin{document}
\maketitle

\begin{abstract}
Mobile robots, performing long-term manipulation activities in human 
environments, have to perceive a wide variety of objects possessing very different visual characteristics and need to reliably keep track of these throughout the execution of a task. In order to be efficient, robot perception capabilities need to go beyond what is currently perceivable and should be able to answer queries about both current and past scenes. In this paper we investigate a perception system for long-term robot manipulation that keeps track of the changing environment and builds a representation of the perceived world. Specifically we introduce an amortized component that spreads perception tasks throughout the execution 
cycle. The resulting query driven perception system asynchronously 
integrates results from logged images into a symbolic and numeric (what we call sub-symbolic) representation that forms the perceptual belief state of the robot. 
\end{abstract}


\section{Introduction}
\label{sec:intro}

Robots are already capable of performing complex 
manipulations tasks, demonstrated through numerous everyday activities such as 
towel folding~\cite{abbeel10icra}, pancake making~\cite{pancakes11humanoids} or 
pipetting in a chemistry lab~\cite{lisca15osd}. None of 
these tasks would be possible without a capable perception framework that is 
able to guide the individual manipulation actions of the robot. Additionally, if we want our robots to alternate between such tasks or perform them repeatedly there is a long-term element to all of them that needs to be taken into account. Robotic agents that are to perform long-term manipulation tasks in realistic environments have to be equipped with sophisticated, robust, and accurate object and scene perception capabilities that go beyond the analysis of individual images.

Consider, for example, a robotic agent that is to perform house chores, such as prepare meals and serve them. To do this it needs to set or clean the table, load and unload the dishwasher, place objects in cupboards, etc. 
The perception system of such a robot, should be able to quickly answer queries like '\textit{where did I see an object out-of place}' or '\textit{is there anything left on the table}' without the extra effort of navigating to an objects location and applying direct perception. In most cases, object perception in robotics is phrased as the problem of detecting a region in the captured sensory data which contains the object the robot asks for. This can be quite a challenging task, as the appearance of objects in human environments can vary tremendously, or the captured RGB-D images might be noisy, the scenes very cluttered with lots of occlusions. The problem gets even worse if the robot has to distinguish the state of objects, that is, when it has to put the clean plates into the cupboard and the dirty ones into the dishwasher. 

\begin{figure}[!t]
\centering
  \includegraphics[width=0.99\columnwidth]{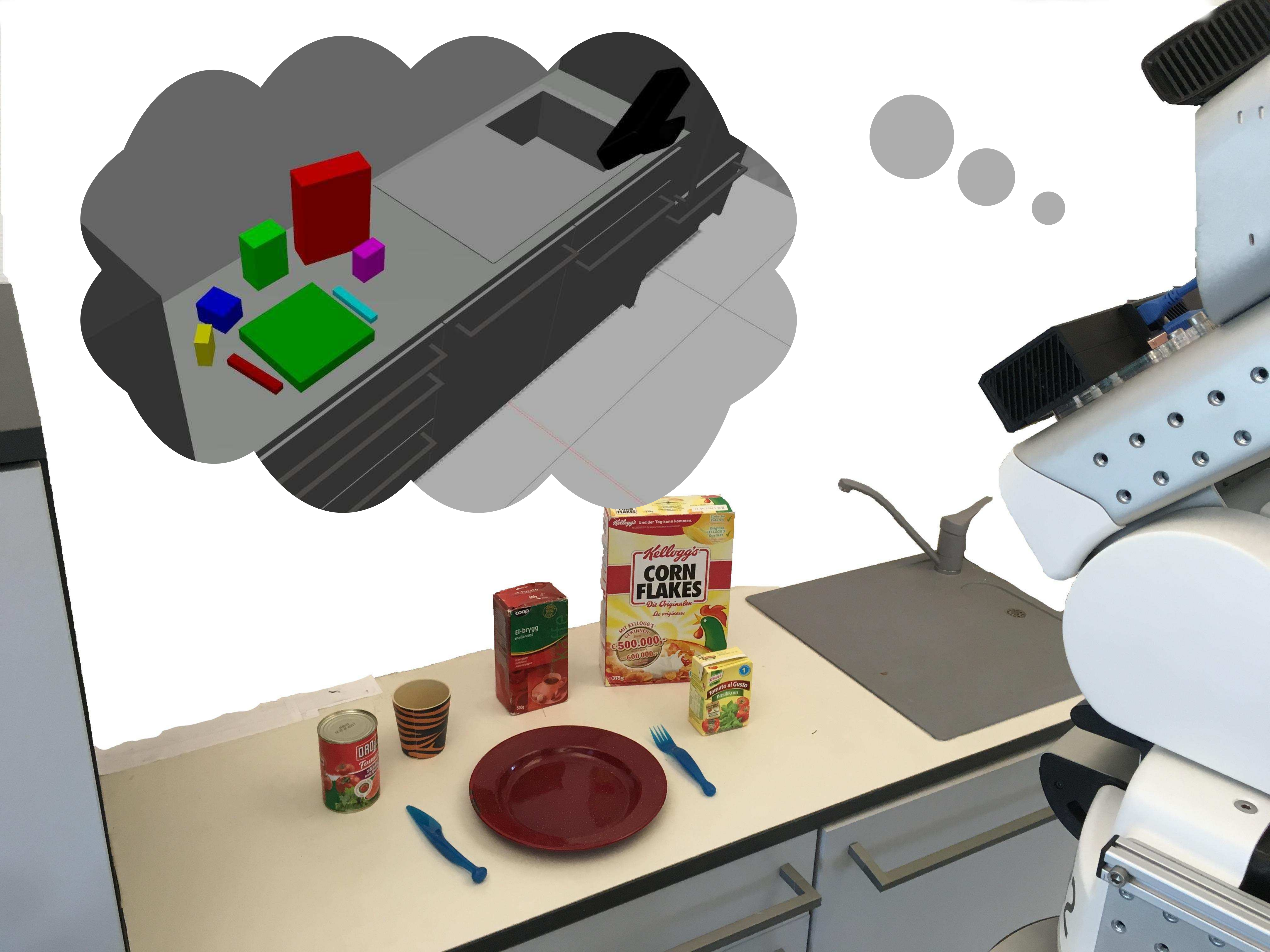}
  \caption{PR2 looking at a table scene, and visualization of its internal 
  representation of the sub-symbolic belief state}
   \label{fig:front_page}
 \vspace{-3ex}
\end{figure}

We propose to phrase robot perception as pervasive query answering which is driven by the tasks and queries formulated using a general query language. We achieve this pervasiveness through an amortized perception system that gradually builds an internal representation of the environment (Figure~\ref{fig:front_page}). In AI and robotics the term \textit{amortized} is most often used in the domain of probabilistic reasoning, as in \textit{amortized inference}, which stands for learning from past inferences, such that future ones run faster. We believe that a robotic agent's perception system needs to behave in a similar fashion. Amortization also refers to spreading an assets cost over it's useful lifetime. To learn from the past inferences we propose to spread the perception tasks over the entire operational lifetime of an agent. This means that perceptual processes should be active even when perception is not directly needed by the robot control system and that the utility of past percepts should be maximized. The central question in implementing the amortization effect is how to maximize the information gain from the logged images such that a higher amount of queries can be answered correctly.

\smallskip

\noindent The core contribution that we propose in this paper is a system that treats  robot perception as a query answering problem and refines beliefs about objects through amortization, separating the generation of symbolic and sub-symbolic beliefs into different computational units and exploiting task and background-knowledge to aid belief state management.

\begin{figure*}[!t]
  \centering
    \includegraphics[width=0.99\textwidth]{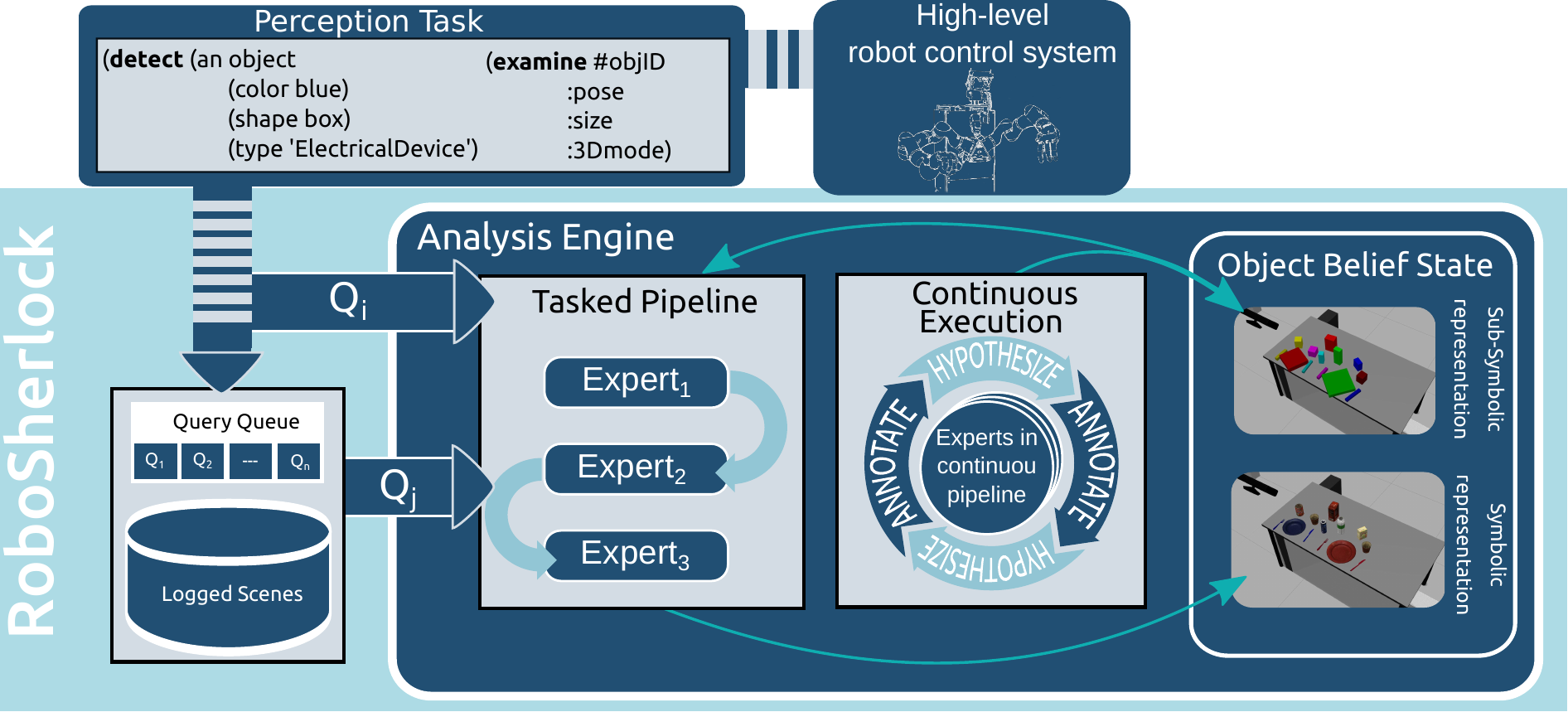}
  \caption{System overview with the interaction of all major components}
  \label{fig:system_overview} 
   \vspace{-3ex}
\end{figure*}

\section{Motivation and Related Work}
\label{sec:motivation}

The proposed amortization effects can only be achieved through the management of a belief-state about the objects a robot encounters. There are several reasons why computing and maintaining a belief state is feasible and can help better inform future perception tasks. First, many environments (such as an apartment, a factory, a surveillance area) change slowly compared to the frequency in which robots capture images of parts of them. Second, during operation, robots typically capture multiple images from different views of the same scenes and therefore can exploit the images from different views to improve results. Third, when the robot acts in the environment it often has spare computational resources, in particular when it navigates from one place to another. In these cases it can perform perception tasks on previously captured images without delaying robot activity. 


The most commonly used perception systems in robotics, as the likes of LineMod~\cite{hinterstoisser13linemod} or more recently deep learning based approaches like YOLO~\cite{redmon2016yolo9000} process images in a one-shot manner. Another category of popular perception systems are ones that enable visual servoing, through the tracking of low-level features in an image (e.g. SimTrack~\cite{pauwels_simtrack_2015}). These tracking systems however, are more oriented towards the short-term tracking of objects during individual manipulation tasks.

The most comprehensive work on object belief states was done by Blodow~\cite{blodow14Phd}, laying the ground work for what is necessary to create and manage such representations. In~\cite{Blodow10Humanoids} Blodow et al. present a Markov Logic Network based framework dynamically resolving the entities of objects. The idea of entity resolution is further developed by Wiedemeyer et al.~\cite{Wiedemeyer15pervasive}, where a pervasive `calm` perception component for the \textsc{RoboSherlock} framework~\cite{beetz15robosherlock}  is built.

Another interesting approach is taken by Lee et al.~\cite{lee14Beliefpropagation}. They argue for the importance of a correct belief management and highlight connections  to biological systems. Milliez et al.~\cite{milliez14roman} and Warnier et al.~\cite{Warnier12ROMAN} present SPARK, a framework for belief management similar to the work presented here, but with a focus more on spatial reasoning and the knowledge component. Parts of SPARK consist of managing an object belief state but since emphasis is 
put on spatial reasoning the scenes and objects used in the experiments are idealized. 

There is an increasing body of literature on semantic mapping and SLAM approaches in robotics, large parts of which are closely related to our work. A thorough survey of these is presented by Cadena et al.~\cite{cadena16SLAM}. In this survey authors identify several open problems, like that of semantic mapping being much more than a categorization problem, leading to a need for high-level, rich representations. Most of the approaches for semantic mapping are concerned with capturing the world around the robot as 
accurate as possible with the agent being an observer rather than an actor in the environment. We see object belief states as something 
complementary to these semantic object maps. Perceptual belief states for manipulation tasks do not have to be one hundred percent accurate since interaction with the world can validate or contradict these beliefs. 

The argument for building belief states in an amortized manner is perhaps best phrased by Gershman et al.~\cite{gershman14amortized} where authors state that the human \textit{``brain operates in the setting of amortized inference, where numerous related queries must be answered [..] in this setting, memoryless algorithms can be computationally wasteful``}. The 
system proposed by us manages and updates beliefs about objects in the environment during the entire execution of a long-term manipulation activity, such as a pick and place tasks. Sub-symbolic and symbolic data is handled by separate computational blocks, but brought together in a unified data structure that serves as the belief state. These two modalities are treated separately since analog to real-time tracking of objects, resolution of object entities is handled by using sub-symbolic information that is available at a higher frequency. Symbolic information about the world can be derived on demand and allows for semantic query-answering.

\section{System Overview}
\label{sec:system}

In order to build the proposed system there are two main components that need to be detailed. First, we explain how symbolic and sub-symbolic information about the objects in the world is generated and stored. Second, we specify how the computational processes that generate this information interact with each other and the robot's high level control system.

%

The central data structure is the \textit{scene}, representing the beliefs of the robot about the environment at a specific timestamp. A scene $S$ is made up of a set of input images $I$ (e.g. depth, RGB), a set of object hypotheses $Hyp$ and a set of scene annotations  (e.g. supporting plane or robot location) $A$. Formally, $S=\langle\{I_{1:l\ }\},\{Hyp_{1:m}\},\{A_{1:n}\}\rangle$, where $Hyp_i$ is an object-hypotheses that consists of a region of interest $Roi$ and a list of key-value pairs $Kvp = \{(k,v)_{1:o}\}$ such that $k \in K$ where $K$ is the set of valid keys and $v$ represents either numerical values or a symbol, depending on properties of $k$. 
\[v= \begin{cases} 
      [v_1,v_2,\ldots,v_l] & v\in \mathbb{R}   \\
      \langle symb \rangle 
   \end{cases}
\]
For example a box-like red object which has a pose and a 3D feature descriptor would be described as:
\begin{equation*}
\begin{split}
\textit{Hyp}_1=\{Roi,[(\textit{shape},box),(\textit{color},red),\\(\textit{pose},[x,y,z,r,p,y]),(\textit{vfh-descriptor},[v_1, \ldots ,v_{307}])]\}
\end{split}
\end{equation*}

\noindent where $Roi$ represents the collection of pixels and/or 3D points that make up the hypotheses in the raw data. 

A long-term robot experiment is thus made up of a chronological 
sequence of scenes and the resulting belief state that is built up over time:  $Exp = ([S_0,S_1, \ldots, S_t],Bs)$ where $t$ represents a globally unique timestamp and the belief state consists of a set of objects $Bs = (O_1,O_2, O_n)$. An object $O_i$ in the belief-state consists of a a time-indexed list of hypotheses that were associated to it. 


Objects that a robot might encounter during the execution of a task can bear various visual characteristics that can only be recognized using different algorithmic approaches. For this reason processing the raw data is tackled in an ensemble of experts approach. This means that processing is split into several special purpose modules (denoted as experts in Figure~\ref{fig:system_overview}) that generate object hypotheses in the RGB-D image captured by the robot or annotate these hypotheses. 

These modules are then grouped into pipelines forming two computational blocks (continuous execution and tasked pipeline in Figure~\ref{fig:system_overview}) that operate in parallel and share a common data structure. The system is driven by the queries formulated by the high-level control system of the robotic agent. A continuous execution block is responsible for generating object hypotheses and the sub-symbolic annotations of these. This same block is also tasked with resolving hypotheses to existing beliefs that make up the sub-symbolic belief state (see Section ~\ref{sec:subsymbolic_bs}). The continuous execution is subject to several task- and background-knowledge based filters (see Section ~\ref{sec:bkg-knowledge} in order to minimize the amount of data needed to be processed. Scenes resulting from images and raw data that gets processed are logged for later processing.  

During the execution of a task, the robot control program sends semantic queries in the form of nested key-value pairs to the system (upper part of Figure~\ref{fig:system_overview}). As described in~\cite{balintbe16task} these queries get interpreted and a tasked pipeline is planned. During execution of this pipeline those objects from the sub-symbolic belief-state that have a matching hypotheses in the current scene refine the beliefs about the objects by updating it with the resulting symbolic information.

At this stage symbolic results get generated only for the objects that are in the scene at the time of the query. This is equal to answering a query through performing one-shot perception. To achieve the desired amortization effect and spread the cost of perception throughout the execution of a task, queries received are buffered and executed on the previously logged images. This is done in parallel and in the background at times when the robot is idling or moving from one place to another. This way the belief-state gets updated with symbolic data that results from object hypotheses found in previous scenes and our representation of objects becomes richer. Queries from the queue are processed in a first come first served manner, beginning with the latest scene. The image filters described in Section~\ref{sec:bkg-knowledge} assure that there is a good degree of data throttling such that we don't waste resources on processing redundant images what would not yield any information gain.

\section {Implementation}

The proposed system was implemented using 
\textsc{RoboSherlock}~\cite{beetz15robosherlock}, an open-source perception framework based on the principles of unstructured information management (UIM)~\cite{ferrucci04uima}. Robot perception in \textsc{RoboSherlock} is treated as a query-answering problem~\cite{balintbe16task}, where task-, domain- and background-knowledge are integrated in the process of generating parametrized pipelines for perception tasks.

Extensions to the framework that enabled taskability~\cite{balintbe16task} and use of background knowledge~\cite{beetz15robosherlock} make the framework ideal for developing our amortization schema on top of it. 

\subsection{Generating sub-symbolic representations}
\label{sec:subsymbolic_bs}

The sub-symbolic representation of objects consists of numerical annotations acquired by the continuous execution component. Initially the system is aware only of hypotheses of objects that have visual descriptors and very low-level geometric information: i.e. estimated pose in the world and oriented 3D bounding box. These low-level percepts, the position in the world and the timestamps help disambiguate between objects of similar characteristics and match hypotheses to existing objects in the belief state. This process is what we call entity resolution. 

The visual features extracted in the continuous execution during the conducted experiments are presented in the upper half of Table ~\ref{tab:lowLvl}.

\begin{table}[ht]
\begin{center}
\begin{tabularx}{\columnwidth}{|l|X|}
    \hline
    \multicolumn{2}{|c|}{Features used for sub-symbolic matching} \\
    \hline\hline
    \textbf{Visual Feature} & Description\\
    \hline\hline
    Colour features & HSV color histogram \\
    \hline	
    3D features & VFH descriptors ~\cite{vfh}\\
    \hline
    Generic descriptors & $DeCAF_{7}$ trained on ImageNet2012~\cite{Donahue_ICML2014}\\
    \hline
    2D-Keypoints/Descriptors & BRISK/FREAK\\
    \hline
    6DOF Pose & 3D estimate from centroids or 2D estimate from Hu-moments\\
    \hline\hline
    \multicolumn{2}{|c|}{Symbolic annotations} \\
    \hline\hline
    color & symbolic color based on HSV color distribution\\
    \hline
    shape & primitive shape, inspired by ~\cite{goron12robotik} \\
    \hline
    class label & k-NN classifier trained on $DeCAF_{7}$ descriptors of partial views of objects from a turn-table\\
    \hline
\end{tabularx}
\end{center}
\caption{Symbolic and sub-symbolic perceptual features and annotators}
\label{tab:lowLvl}
\vspace{-3ex}
\end{table}

In order to solve entity resolution we define a distance function $dist_{k}$ for every key $k \in K$. Each function maps the distance between the values associated with $k$ to $[0\ldots1]$.

$$dist_k(k_1, k_2) \rightarrow [0\ldots1]$$

For the sub-symbolic annotations used in this work, this distance function is defined based on the Euclidean distance between the values. Symbolic annotation are not taken into account during entity resolution. For example the distance function between two poses is defined as:

$$dist_{pose}(p_1,p_2) = min(1, 4*\|p_1-p_1\|))$$

For each percept a weight $w_k$  is chosen representing its significance. In order to quantify the similarity between a object hypotheses seen in the current scene $Hyp_m$ and an object from the belief state $O_1$  the function $sim(Hyp_m,O_1)$ is defined which is the normalized weighted sum of all distances for all pairs of low-level annotations $S_k$ and $B_k$:

$$sim(Hyp_m,O_1) = 1 - {\sum\limits_{k\in K} w_k * dist_k(S_k, B_k) \over \sum\limits_{k\in K}w_k} \rightarrow [0\ldots1]$$

The resolution of identities is done by comparing descriptions of object hypotheses in the current scene, $\langle Hyp_i \rangle$, with the description of objects from the belief state $\langle Obj_j \rangle$. A fast-matching between objects in the belief state that had been seen in the previous scene, and should still be in view, is performed with the object hypotheses from the current scene. The fast-match quickly finds candidate pairs, based on the distance between the known and perceived object pose and the time that had passed since it was last seen. For each pair the similarity is computed and if it exceeds a certain threshold, the two get merged. This speeds up the validation process of cases when the robot simply moves to look at the same thing from a different angle. 

Next, for all remaining pairs of object hypotheses and objects $\langle  Hyp_i,Obj_j \rangle$ a probability is calculated that they depict the same object in the real world using the previously introduced similarity function. The most probable solution, given that it exceeds a certain threshold, is chosen and the hypotheses gets added to the existing object, otherwise a new object is created and added to the belief state.

\subsection{Symbolic belief management and amortization}
\label{sec:symbolic_bs}

\begin{figure}[!ht]
  \centering   
  \includegraphics[width=0.99\columnwidth]{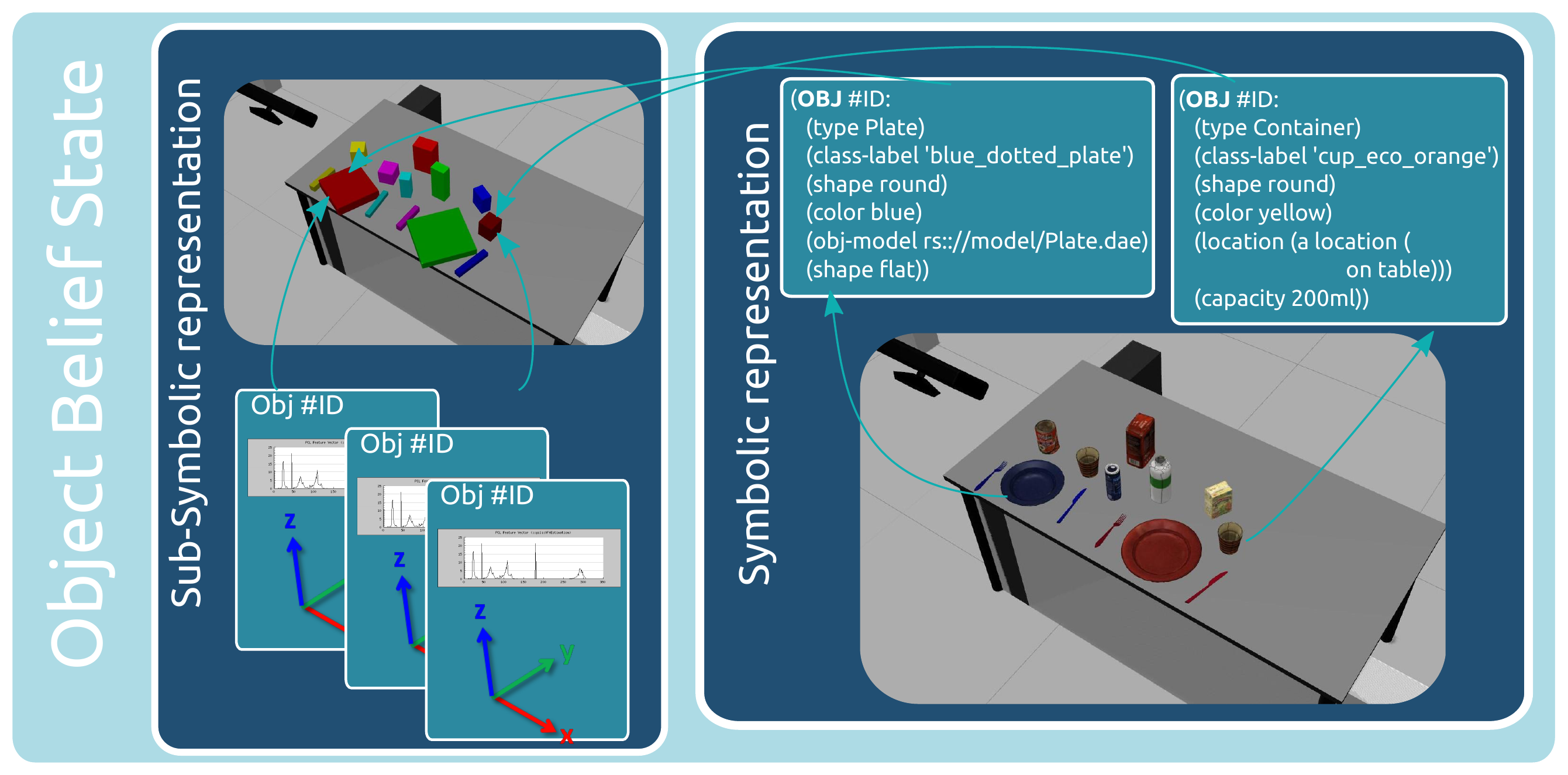} 
  \caption{Evolution of the belief state, as queries get answered.}
  \label{fig:representation}
\end{figure}
  
In order to be competent at answering complicated queries, robots need to have semantic information about the objects in the environment. Most of the state of the art algorithms used in object recognition and scene understanding, are not able to handle the (close-to) real-time requirements of mobile robots, or 
if they do, they only handle a subset of the objects in the environment (e.g. SimTrack for textured objects). Because of this, symbolic data in the system is introduced through asynchronous query answering. Queries are formed using the meta-language described in~\cite{balintbe16task}. Similarly to the representation of an object hypotheses $Hyp$ the queries consist of a list of key-value pairs. A very simple query is shown below, asking for a flat shaped object that is black:

\smallskip
\begin{small}
\begin{tabbing}
  (detect (an \= object \\
  \> (shape flat) \\
  \> (color black)))
\end{tabbing}
\end{small}

\noindent The symbolic information maintained in the belief state depends on the queries that the high-level control system of the robot issues, since they only get generated for the scenes that are being processed during query answering. The right side of Figure~\ref{fig:representation} depicts the link between the symbolic and sub-symbolic beliefs. Additional to the low-level percepts,  symbolic information, as the likes of color, shape or class label, are linked to these beliefs.

The sources of symbols in our system are presented in the bottom half of Table~\ref{tab:lowLvl}. Handling contradictory symbolic data of the same type has been detailed in our previous work ~\cite{icra14ensmln}, where a first order probabilistic reasoning system based on Markov Logic Networks was used. In this work we only use one source of information for each type of symbol (shape, color, type), and focus on how these contradictions affect query answering and to what extent can the be remedied using amortization.
For each object in the belief state we maintain a history of the resulting symbolic values from hypotheses that were associated to it. Although in this work they are treated as simple string entities, the symbols generated are grounded in a Knowledge base, allowing users of the system to perform knowledge-based reasoning on them. The values of these symbols range from shape characteristics (box, round etc. ) to object class names. 

\subsection{Integrating background and task-knowledge}
\label{sec:bkg-knowledge}
%

%
%

Background- and task-knowledge offer valuable sources of information that ensure the feasibility of processing logged data during task executions. For example, given a semantic map of the environment~\cite{iros12semantic_mapping} 
and a localized robot, it is trivial to filter out any parts of the camera images that are not in scope of the current task, e.g. in a pick and place task only the source and destination regions are of interest. This reduces false detection and computational effort.

Besides this in many robotic tasks the scenes are mostly static and successive frames are often similar and lead to no information gain, therefore skipping them offers more processing time for other tasks. Another common source for erroneous detections are motion blurred images. This happens if the camera or something in view is moving fast (like the manipulators of the robot). For the camera movement the pose of the camera in world space is tracked and movements bigger than a certain threshold are not considered. For the motion blur we use the variation of the Laplacian detailed by Pech-Pacheco et. al~\cite{blur@icpr2000}. Using a central knowledge-base, the perception system can also query for the task-knowledge asserted by the high-level plans, in order to find out if the robot will move over longer distances without needing object perception. Taking advantage of this knowledge saves us the trouble of processing images that would have no meaning for the task that needs execution.


\section{Experimental Analysis}
\label{sec:experiments}

The goal of the amortization effect is to maximize the information gain from the logged images such that a higher amount of queries can be answered correctly.  Answering a query correctly at any given time during a task is equivalent to having the correct symbolic representations in the belief-state at any given time.  Long-term tasks require that the performance does not degenerate over time so besides the correctness of the symbolic annotations we analyze how different time windows for amortization affect the end results, through the introduction of an amortization coefficient. 

\begin{table}[t]
	\begin{center}
		\begin{tabular}{|l|l|l|l|l|l|l|l|l|}
			\hline
			& \multicolumn{2}{c|}{Ep.1} & \multicolumn{2}{c|}{Ep.2}& \multicolumn{2}{c|}{Ep.3}& \multicolumn{2}{c|}{Ep.4}\\
			\hline
			\# of Objs.  &\multicolumn{2}{c|}{9} & \multicolumn{2}{c|}{15}& \multicolumn{2}{c|}{20}& \multicolumn{2}{c|}{25}\\
			\hline
			duration &\multicolumn{2}{c|}{277(s)} & \multicolumn{2}{c|}{328(s)}& 
			\multicolumn{2}{c|}{510(s)}& \multicolumn{2}{c|}{520(s)}\\
			\hline
			\# of $Hyp$.  &\multicolumn{2}{c|}{130} & \multicolumn{2}{c|}{353}& \multicolumn{2}{c|}{694}& \multicolumn{2}{c|}{759}\\
			\hline
			\# pnp tasks &\multicolumn{2}{c|}{3} & \multicolumn{2}{c|}{4}& 
			\multicolumn{2}{c|}{6}& \multicolumn{2}{c|}{10}\\
			\hline
			filters & Off & On& Off & On& Off & On& Off & On\\
			\# objs. in bs.  & 15 & 9& 26 & 20& 49 & 28& 54 & 31\\
			\hline				  
		\end{tabular}
	\end{center}
	\caption{Summary of episodes of the conducted experiments}
	\label{tab:episode_details}
\end{table}

\begin{figure}[]
	
	\includegraphics[width=0.24\textwidth]{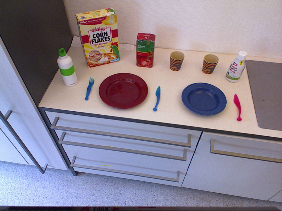}
	\includegraphics[width=0.24\textwidth]{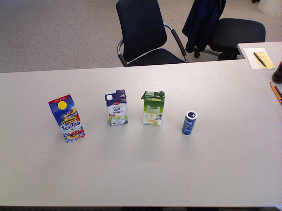}
	\caption{Initial state of the two table top scenes in episode 2. }
	\label{fig:ep_scenes} 
\end{figure}

The experiment were conducted using four different episodes with increasing difficulty, where the task of the robot is to move objects from one supporting surface to another. Example of a scene from one of the episodes is shown in Figure~\ref{fig:ep_scenes}. Each image of the episodes was hand-labeled to contain ground-truth about the relevant symbolic information of the objects (shape, color and class).

In all cases the experiments were conducted using a localized\footnote{we use adaptive Monte-Carlo localization (\url{wiki.ros.org/amcl})} PR2 robot that was performing pick-and-place tasks in a kitchen environment\footnote{Actual picking and placing of the objects was handled by a human, in order to reduce the complexity}. The 25 objects chosen for the experiment are typical house-hold items that posses varied visual characteristics. Some are flat and textureless while others are textured and well visible. The objects are clearly separated with few occlusions, since we are only interested in analyzing the amortization effects of the annotations. We believe that the same principles of amortization could be applied also for the case of hypotheses generation in challenging cluttered, occluded scenes and investigating these would merit a separate endeavor. The hypotheses are generated using a combination of 3D Euclidean clustering using the table top assumption combined with a threshold based binary color segmentation useful for small flat objects that otherwise would not be found. For the classification we used k-NN detailed in Table~\ref{tab:lowLvl}, trained on the 25 objects from episode 4, which subsumes the the objects used in the other three episodes.  Details of the four runs such as total duration of an episode, number of pick and place tasks (\# pnp tasks), number of objects  at the end of execution in the belief state (\# objs. in bs.) and number of total object hypotheses in the episode (\# of $Hyp$) are presented in Table~\ref{tab:episode_details}.

Since the performance of the system is dependent n the performance of its components we start by measuring a baseline performance of the system . We first look at the correctness of the sub-symbolic belief-state and the effects of using the background- and task-knowledge enabled filters. We continue  with a qualitative analysis of amortization, followed by a quantitative analysis of the system. In all cases the baseline that we compare against is the performance of one-shot single scene interpretation. 

\subsection{Sub-symbolic management of the belief state}

To perform the evaluations the low-level pipeline and the identity resolution of the objects, we follow the experimental set-up described in~\cite{Wiedemeyer15pervasive}.

\begin{table*}[t]
	\begin{center}
		\begin{tabular}{|l|r||c|c|c||c|c|c||c|c|c||c|c|c||c|c|c||}
			\hline
			&  & \multicolumn{3}{c||}{\textbf{Ep.1}} & \multicolumn{3}{c||}{\textbf{Ep.2}}& \multicolumn{3}{c||}{\textbf{Ep.3}}& \multicolumn{3}{c||}{\textbf{Ep.4}} & \multicolumn{3}{c||}{\textbf{Average}}\\
			\hline
			& & \textbf{A}& \textbf{P}& \textbf{R} & \textbf{A}& \textbf{P} & \textbf{R} & \textbf{A}& \textbf{P}& \textbf{R} & \textbf{A}& \textbf{P}& \textbf{R}& \textbf{A}& \textbf{P}& \textbf{R} \\
			\hline
			\multirow{ 2}{*}{Shape} & One-shot & 0.68& 0.82& 0.69 & 0.65& 0.67& 0.65& 0.66& 0.67& 0.66& 0.63& 0.64& 0.64& 0.65& 0.7& 0.66\\ 
			& Amortized & 0.76 & 0.83& 0.76 & 0.68& 0.71& 0.69& 0.73& 0.75& 0.73& 0.68& 0.68& 0.68& 0.71& 0.74& 0.71\\ \hline
			\multirow{ 2}{*}{Color} & One-shot & 0.87& 1.0& 0.87 & 0.91& 0.95& 0.92& 0.82& 0.9& 0.83& 0.84& 0.92&0.84& 0.86&0.94&0.86\\ 
			& Amortized & 0.89& 1.0& 0.89 & 0.98& 0.99& 0.99& 0.86& 0.93& 0.87& 0.85& 0.91& 0.85& 0.89& 0.95& 0.9\\ \hline
			\multirow{ 2}{*}{Class} & One-shot & 0.93& 0.95& 0.93 & 0.98& 0.98& 0.98& 0.95& 0.96& 0.95& 0.96& 0.97&0.96& 0.96& 0.96&0.96\\ 
			& Amortized &0.99& 0.99& 0.99&0.96& 0.96& 0.96&0.92&0.94&0.92& 0.89&0.90&0.90&0.92&0.92&0.93\\ \hline\hline
			\multicolumn{14}{r||}{Coverage one-shot}& \multicolumn{3}{c||}{82.2 \%} \\
			\cline{15-17}
     		\multicolumn{14}{r||}{Coverage amortized}&  
     		\multicolumn{3}{c||}{94.3 \%} \\
     		\cline{15-17}
		\end{tabular}
	\end{center}
	\caption{Accuracy (\textbf{A}), Precision(\textbf{P}) and Recall(\textbf{R}) for shape, color and class annotations of object hypotheses with and without the amortization effects.}
	\label{tab:experiment_details}
    \vspace{-3ex}
\end{table*}

 
The number of actual objects in the environment compared with the number of objects in the belief-state at the end of a pick and place task give us an insight into how well we are able to track objects over time. It is important that the belief state reflects the reality as close as possible since it affects the number of results generated during query answering. We also highlight the usefulness of the task- and background-knowledge by running each episode with and without the filters in place (row six of Table~\ref{tab:episode_details}). We observe a significant improvement of the belief state when running with the filters turned on. This improvement is mainly due to the skipping of motion blurred images, which have a negative impact on hypotheses generation. For analyzing the effects of amortization we thus make use of these filters.

\subsection{Benefits of amortization}
\label{sec:benefits}

Amortization is the integration of results from past queries executed on logged scenes into the belief-state of the robot. Answering a query correctly, means having the correct symbolic data deduced from the current scene. Quantifying the effects of amortization thus means comparing results of scenes in isolation (equivalent of one-shot perception) with the results of the same scenes when integrating results from the past.

\begin{table}[]
	\begin{center}
		\begin{tabular}{c|c|c|c|c}
			
			id &Object & \# obj.hyp &  \#classifications & \% correct \\
			\hline\hline				 
			\rowcolor{LightCyan}
			3 & blue knife  & 127 & 81  & 30.8 [\%]  \\
			
			5 & red plate   & 83  & 81  & 97.5 [\%]  \\
			\rowcolor{gray}
			8 & sigg bottle & 83  & 7   & 100  [\%]  \\
			
			\rowcolor{gray}
			10& yogurt      & 62  & 25  & 100  [\%]  \\
			11& cup         & 75  & 60  & 86.6 [\%]  \\
			\rowcolor{gray}
			12& ice tea     & 176 & 51  & 100  [\%]  \\
			\rowcolor{LightCyan}
			13& soja milk   & 176 & 176 & 57.9 [\%]  \\
			
			\rowcolor{gray}
			15& salt        & 176 & 30  & 100  [\%]  \\
			16& blue knife  & 90  & 85  & 83.5 [\%]  \\

		\end{tabular}
	\end{center}
	\caption{Analysis of object hypotheses per object in belief state. Gray rows: few successful classifications; Cyan rows: low precision}
	\label{tab:hypotheses_analysis_per_obj}
	\vspace{-4ex}
\end{table}

We start our analysis by first looking at a qualitative analysis of classification results on representative objects from Episode 2. For the base-line performance the confidence threshold (one minus the distance of the normalized descriptors) of the k-NN is set to 0.6.

A query for objects in scenes when their hypotheses is misclassified or not classified at all would not result in a correct answer. The color coding in Table~\ref{tab:hypotheses_analysis_per_obj} highlights two distinct situations that largely contribute to wrong answers of this kind. The case of the ``sigg bottle``, ``salt`` and ``ice tea`` are worth highlighting. All three objects have a high number of hypotheses attached to them, but a low number of classification results, all of which are correct. 

\begin{figure}[b]
  
\includegraphics[width=0.5\textwidth]{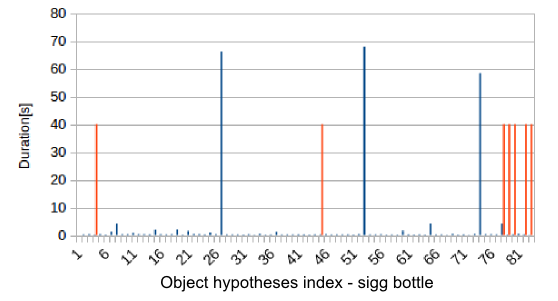}
  \caption{Results of classification for the hypotheses of the \textit{sigg bottle}. Blue bars 
  represent the time elapsed between hypotheses generation, red bars mark the 
  hypotheses that are correctly classified. Amplitude of red bars is only for ease of viewing. x-axes represents hypotheses of the object ordered chronologically.}
  \label{fig:analysis}
\end{figure}

\begin{figure*}
	\includegraphics[width=0.245\textwidth]{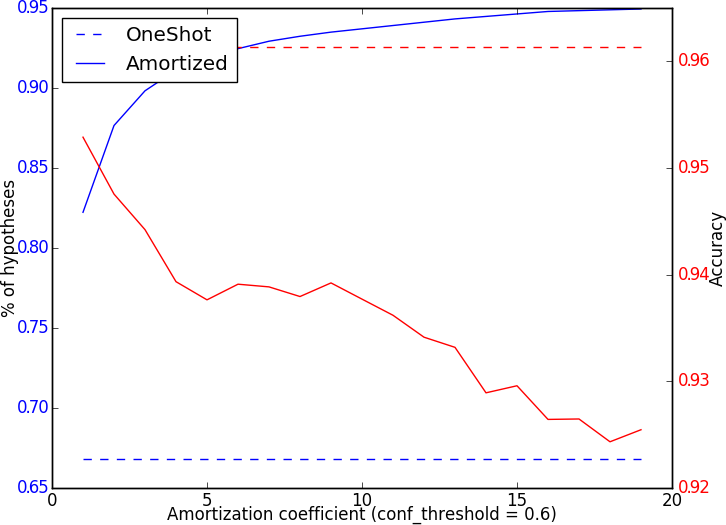}
	\includegraphics[width=0.245\textwidth]{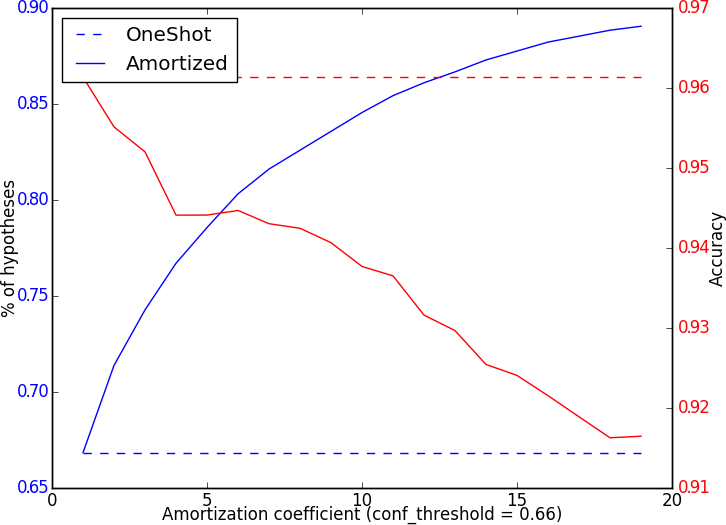}
	\includegraphics[width=0.245\textwidth]{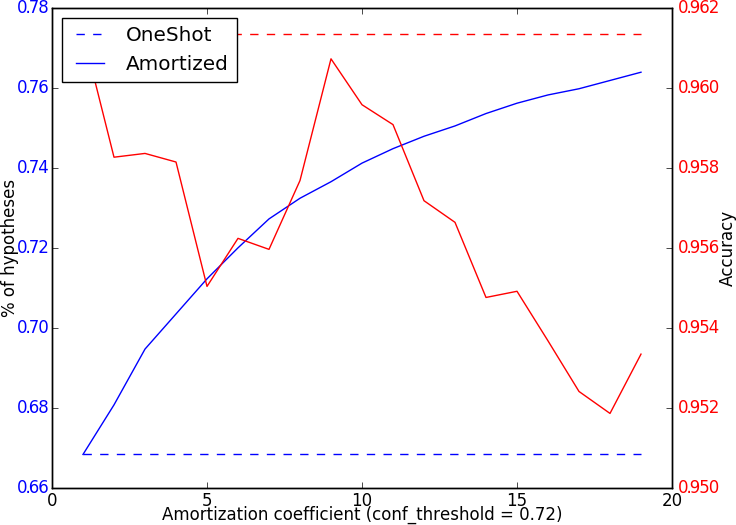}
	\includegraphics[width=0.245\textwidth]{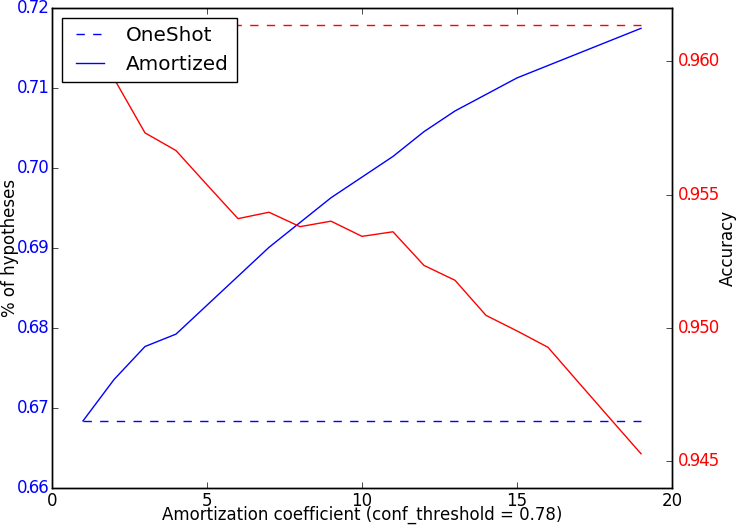}
	\caption{Trade-offs between hypotheses coverage and accuracy when varying confidence thresholds and the amortization coefficient. Dotted horizontal lines represent the baseline results of one-shot perception. These are not affected by the amortization coefficient. Blue lines represent the coverage, red lines the average accuracy. Reported results are an average over all four episodes.}
			\label{fig:analysis}
\end{figure*}

\begin{figure}
	\includegraphics[width=\columnwidth]{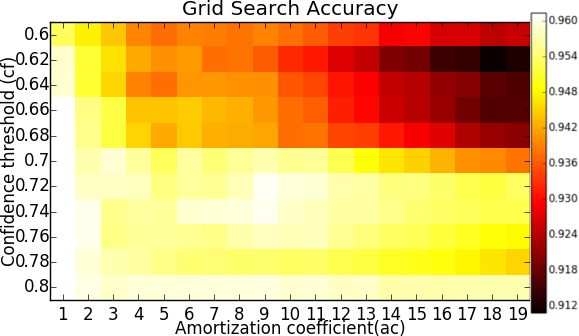}\\
	\\
	\includegraphics[width=\columnwidth]{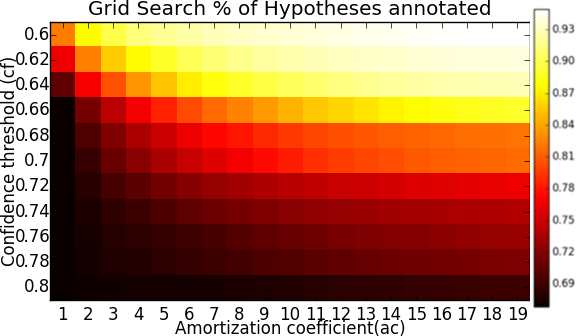}
	\caption{Heat maps of gird search performed on the tuning parameters}
	\label{fig:heatmaps}
\end{figure}

For the effects of amortization to take effect two thing need to happen: there needs to be a query in the buffer that solicits a symbolic result and enough time needs to pass such that the buffered query can be executed on a logged image where the symbol can be correctly detected. In Figure~\ref{fig:analysis} the blue bars are the duration between two consecutive hypotheses in seconds. Whenever we have a large time gap, the robot is either moving or one of the filters is active, allowing for the buffered queries to execute. Between the two hypotheses that match the object at $Hyp_26$ and $Hyp_27$ there is a time gap of more than 60 seconds. If in this time we manage to process enough scenes to reach the one of $Hyp_4$ (where the red bar marks a successful classification), all future queries for the ``sigg-bottle`` could be successfully answered.
 

To quantify the effect of amortization we look at how it affects the symbols deduced for the object from the belief state at every scene from all four episodes. Table~\ref{tab:experiment_details} reports the performance measures of one-shot perception for each three symbols (shape, color, class-label) separately.  The object class results represent an interesting scenario. Even though the average accuracy and precision of the classification is high, only 82.2~\% of hypotheses generated in the four episodes are annotated. We refer to this as the coverage of the results. Thus, the confidence of the k-NN classifier for a large proportion of the object hypotheses does not meet the classification threshold. Accuracy and coverage of classification results are thus our two main performance measures when analyzing the effects of amortization. 

The amortization process depends on two main parameters: 
\begin{itemize}
	\item an \textit{amortization coefficient ($ac$)}: defining how far into the past do we want to go when integrating results from past hypotheses; This coefficient is expressed in occurrences of object in past scenes rather then absolute time, balancing out the frequency at which objects appear in scenes; $ac = 12$ means integrate the results from the previous 12 occurrences (hypotheses) of an object
	\item and a \textit{confidence threshold ($cf$)}: defining how confident the result from the past hypotheses have to be to be taken into consideration
\end{itemize}

A larger amortization coefficient allows the integration of more results from the past (i.e. better coverage) at the cost of classification performance (lower accuracy). In turn, higher confidence thresholds yields higher average accuracy at the cost of coverage. This trend is clearly visible in Figure~\ref{fig:analysis} which illustrates the inverse proportional relation of the accuracy and coverage with $cf= {0.6,0.66,0.72,0.8}$ and $ac = {1:20}$.
 
\begin{figure}[t]
	\includegraphics[width=1.0\columnwidth]{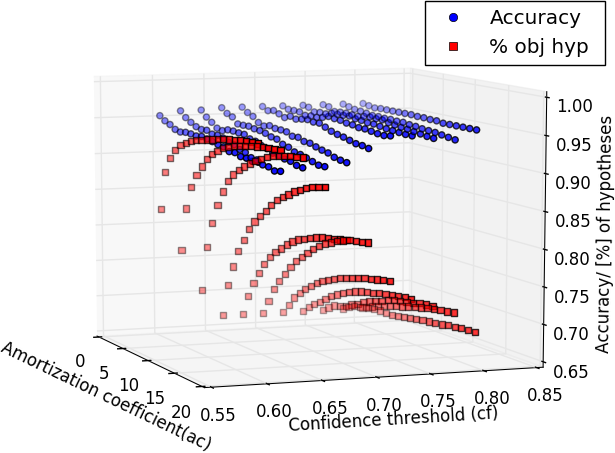}
	\caption{Relationship of coverage, accuracy of class annotation, confidence threshold and amortization coefficients}
	\label{fig:scatter}
\end{figure}

In order to find the best combination of the parameters we perform a grid search, results of which are visualized in the heat maps of Figure ~\ref{fig:heatmaps}. The best choice of parameters is along the line of intersection of these two manifolds given by $max(ac +cf)$ (Figure~\ref{fig:scatter}) and is found to be at $cf = 0.62$ and at $af = 12$. In practice any value around these yields satisfactory results, the differences in performance being minor. Table~\ref{tab:experiment_details} presents the performance measures on the four episodes using these parameters. Overall, in the case of shape and color we can notice a general increase in all metrics, while in the case of classification there is a slight decrease in accuracy and precision but with a 10 \% increase in coverage. Since the accuracy, precision or recall changes of the k-NN are in the range of a couple of percent, it is a better choice to select parameters where these measures are slightly worse but the coverage increases as much as possible. 

Considering that queries can contain more than one key at a time, the increase of the number of queries that can be correctly answered using amortization is considerably larger.

\section{Discussion}
\label{sec:discussions}

We have presented a system that spreads the perception task throughout the lifetime of a robotic agent and maintains a joint symbolic and sub-symbolic 
belief state, for a mobile robot performing pick and place tasks in a human
environment. This enables us to treat perception in a more pervasive manner. We have shown that using available task and background knowledge
a dynamic belief state can be correctly managed in a timely manner, without hindering robot action. The benefits of our approach lie in the way we extend the semantic query capabilities of the robot to answer questions not only about what it perceives in the current scene but also about past percepts.

In the current implementation the system uses its filters to decide when to run pervasive querying on the logged images (e.g. when several consecutive frames are not processed). Furthermore there is no ranking of the queries added to the queue. As future prospect we are planning using episodic memories of performed tasks in order to estimate how much time certain robot operations might take in order to better manage our resources, and schedule tasks in a less invasive manner. We will also investigate ways of ranking queries (e.g. based on which information gain) allowing for sorting and merging of pending perception tasks.



\bibliographystyle{IEEEtran}
\bibliography{literature}

\end{document}